\documentclass[journal]{IEEEtran}

\usepackage{amsmath,graphicx,mathrsfs}
\usepackage{bbm}
\usepackage{etex}
\usepackage{float,subfigure}
\usepackage{cite}
\usepackage{color}
\usepackage{hyperref}
\usepackage{fontawesome}
\usepackage{cite}
\usepackage{amsmath,amssymb,amsfonts}
\usepackage{algorithmic}
\usepackage{graphicx}
\usepackage{textcomp}
\usepackage{booktabs}
\usepackage{multirow}
\usepackage{footnote}
\usepackage{stfloats}
\usepackage[dvipsnames]{xcolor}
\usepackage{arydshln} 
\hypersetup{
	colorlinks=true,
	linkcolor=red,
	filecolor=blue,      
	urlcolor=blue,
	citecolor=blue,
}

\newcommand{\tabincell}[2]{\begin{tabular}{@{}#1@{}}#2\end{tabular}}

\ifCLASSINFOpdf
\else
\fi

\begin{document}
	
	\title{Regression-free Blind Image Quality Assessment with Content-Distortion Consistency}
	
	\author{Xiaoqi Wang,~\IEEEmembership{Student Member,~IEEE},
		Jian Xiong,~\IEEEmembership{Member,~IEEE}, Hao Gao,~\IEEEmembership{Member,~IEEE},  Yun Zhang,~\IEEEmembership{Senior Member,~IEEE},
		and Weisi~Lin,~\IEEEmembership{Fellow,~IEEE}
		
\thanks{Xiaoqi Wang and Yun Zhang are with the School of Electronics and Communication Engineering, Sun Yat-sen University, Shenzhen, 518107, China (e-mail: \href{mailto:wangxq79@mail2.sysu.edu.cn}{\textcolor{black}{wangxq79@mail2.sysu.edu.cn}}; \href{mailto:zhangyun2@mail.sysu.edu.cn}{\textcolor{black}{zhangyun2@mail.sysu.edu.cn}}).~(\emph{Corresponding author: Yun Zhang}.)} 
\thanks{Jian Xiong is with the College of Telecommunications and Information Engineering, Nanjing University of Posts and Telecommunications, Nanjing 210003, China (e-mail: \href{mailto:jxiong@njupt.edu.cn}{\textcolor{black}{jxiong@njupt.edu.cn}}).}
\thanks{Hao Gao is with the College of Artificial Intelligence, Nanjing University of Posts and Telecommunications, Nanjing 210003, China (e-mail: \href{mailto:tsgaohao@gmail.com}{\textcolor{black}{tsgaohao@gmail.com}}).}
\thanks{Weisi~Lin is with the School of Computer Science and Engineering, Nanyang Technological University, Singapore, 639798 (e-mail: \href{mailto:wslin@ntu.edu.sg}{\textcolor{black}{wslin@ntu.edu.sg}}).}
	}
	\markboth{~Vol.~14, No.~8, August~2023}%
	{Shell \MakeLowercase{\textit{et al.}}: Bare Demo of IEEEtran.cls for IEEE Journals}
	
	\maketitle

\begin{abstract}
	The optimization objective of regression-based blind image quality assessment (IQA) models is to minimize the mean prediction error across the training dataset, which can lead to biased parameter estimation due to potential training data biases. To mitigate this issue, we propose a regression-free framework for image quality evaluation, which is based upon retrieving locally similar instances by incorporating semantic and distortion feature spaces. The approach is motivated by the observation that the human visual system (HVS) exhibits analogous perceptual responses to semantically similar image contents impaired by identical distortions, which we term as content-distortion consistency. The proposed method constructs a hierarchical \textbf{$k$}-nearest neighbor~(\textbf{$k$}-NN) algorithm for instance retrieval through two classification modules: semantic classification (SC) module and distortion classification (DC) module. Given a test image and an IQA database, the SC module retrieves multiple pristine images semantically similar to the test image. The DC module then retrieves instances based on distortion similarity from the distorted images that correspond to each retrieved pristine image. Finally, quality prediction is obtained by aggregating the subjective scores of the retrieved instances. Without training on subjective quality scores, the proposed regression-free method achieves competitive, even superior performance compared to state-of-the-art regression-based methods on authentic and synthetic distortion IQA benchmarks. The codes and models are available at~\href{https://github.com/XiaoqiWang/regression-free-iqa}{\color{VioletRed}{\textbf{https://github.com/XiaoqiWang/regression-free-iqa}}}.

%
\end{abstract}
	

	\IEEEpeerreviewmaketitle
	
\section{Introduction}
	\IEEEPARstart{O}{bjective} image quality assessment (IQA) studies algorithms that automatically evaluate the visual quality of images as human judgments. It has been extensively utilized in various applications, including benchmarking visual tasks such as image restoration \cite{restoreiccv,Controiccv}, image compression \cite{comiccv,modalitycompression}, and super-resolution \cite{supericcv,low_resolution}, as well as for quality monitoring in various systems \cite{waterIiqa, chow2016review}. Blind IQA (BIQA) has gained popularity in the IQA community \cite{diivine,moderniqa} over the past decade as a method for evaluating image quality without requiring reference information. 
	
	In developing BIQA models, two main categories of distorted images are considered: synthetic distorted images and authentic distorted images. Synthetic distorted images are created in a controlled laboratory setting by degrading pristine images with a series of types and intensities of distortion \cite{csiq,tid2013,kadid-10k,livemd}. On the other hand, authentically distorted images originate from real-world scenarios and contain naturally occurring distortions \cite{livec, koniq-10k}. Furthermore, researchers have compiled datasets of algorithmically distorted images generated by image processing operations~\cite{pipal}, e.g., denoising and restoration. Then, each distorted image is evaluated by many human observers and assigned an average subjective quality score, namely mean opinion score (MOS or DMOS). A regression-based BIQA model is obtained by fitting the predicted quality scores of training images to their MOS values. 
	\begin{figure}[!tbp]	
		\centering
			\includegraphics[width=\linewidth]{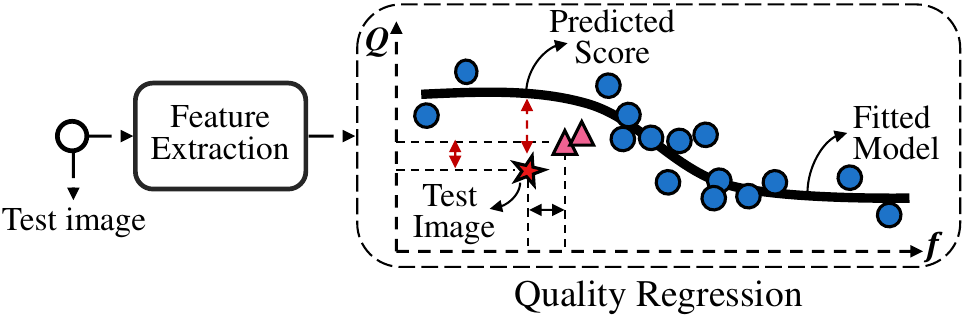}\noindent \vspace{-1mm}
		
		\caption{Overview of regression-based BIQA model. The vertical coordinates represent the image quality~$\textbf{\textit{Q}}$, and the horizontal coordinates represent the feature space~$\textbf{\textit{f}}$ simplified as a one-dimensional space. Given a test image (red star), when the training data distribution exhibits bias,  the regression model may yield biased parameter estimates, leading to substantial prediction error. The image quality can be evaluated by retrieving similar instances in the feature space (pink triangles) to avoid dependence on model parameters.} 
		\label{fig_intro}
	\end{figure}
	
	Regression-based BIQA models are composed of two main components: quality-aware feature extraction and image quality regression, as illustrated in Fig.~\ref{fig_intro}. In traditional models~\cite{nrsl,brisque,diivine,bliinds-ii,cornia,hosa,qac}, researchers have mainly utilized hand-crafted features, e.g., locally normalized luminance coefficients~\cite{brisque}, gradient magnitude and Laplacian of Gaussian response~\cite{gmlog}, to extract quality-aware features. Then, these features are regressed to subjective scores by a nonlinear function, e.g., support vector regression (SVR)~\cite{brisque,cornia,nrsl}. Due to the limitations of hand-crafted features and regression functions, recent years have seen increasing use of deep learning-based BIQA models. In deep learning-based models~\cite{biecon,rankiqa,meon,dbcnn,meta,clriqa,sgdnet,hyperiqa,tres}, convolutional neural networks (CNN) are usually employed for extracting quality-aware features, and then the image quality score is obtained by a quality regressor, e.g., multi-layer perceptron (MLP)~\cite{deeprn,sgdnet}, hyper network \cite{hyperiqa}, Transformer encoder \cite{tres,triq}.
	
	Nevertheless, regression-based BIQA models are susceptible to the representativeness and frequency of training samples. Specifically, the size-limited IQA databases can result in biased training samples, compromising the accuracy of modeling the true distribution of image qualities. Moreover, the regression-based models are derived by minimizing the average prediction error of all training samples, with emphasis on frequent ones. These factors may contribute to a biased estimation of the model parameters~\cite{gelman2006bias}, causing inaccurate prediction. As shown in Fig.~\ref{fig_intro}, the vertical and horizontal coordinates represent image quality and features, respectively. For clarity, the images are shown as points in a one-dimensional space and the black curve represents the fitted BIQA model. If the training set has an inadequate representation of the real sample distribution, the model parameters estimated by the regression loss functions can introduce substantial prediction bias for the test samples, such as the red star in the figure. To alleviate this issue, it is important to avoid over-reliance on specific model parameters. A possible solution is to assess the image quality by retrieving similar instances, such as the pink triangles, in the feature space.

	\begin{figure}[tbp!]
		\centering
		\subfigure[ ]{%
			\begin{minipage}[t]{0.49\linewidth}
				\centering
				\includegraphics[width=\linewidth]{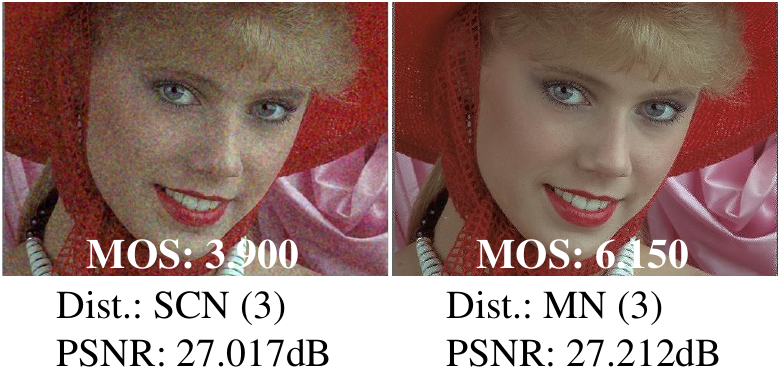}\noindent\vspace{-1mm}
				\label{fig:fig2-a}
			\end{minipage}%
		}%
		\subfigure[ ]{%
			\begin{minipage}[t]{0.49\linewidth}
				\centering
				\includegraphics[width=\linewidth]{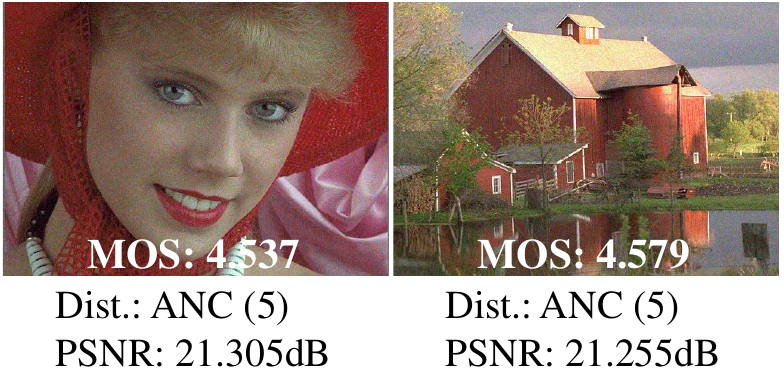}\vspace{-1mm}
				\label{fig:fig2-b}
			\end{minipage}%
		}%
		
		\subfigure[ ]{%
			\begin{minipage}[t]{0.49\linewidth}
				\centering
				\includegraphics[width=\linewidth]{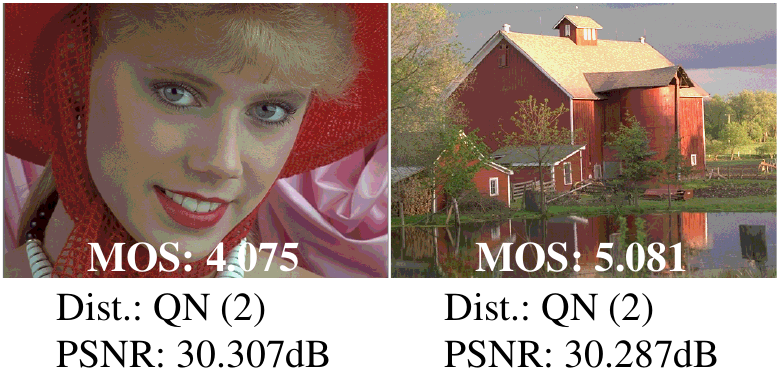}\noindent\vspace{-1mm}
				\label{fig:fig2-c}
			\end{minipage}
		}%
		\subfigure[ ]{%
			\begin{minipage}[t]{0.49\linewidth}
				\centering
				\includegraphics[width=\linewidth]{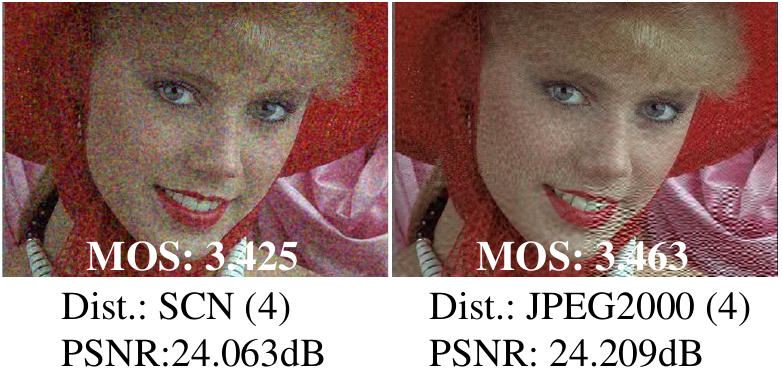}\vspace{-1mm}
				\label{fig:fig2-d}
			\end{minipage}
		}%
		\centering
		\caption{Illustrations of image quality scores not accurately reflecting image content and distortion in TID2013~\cite{tid2013}. The abbreviation "Dist." stands for image distortion, including the type (and level) of distortion.}
		\label{fig_dist_cont}
	\end{figure}
	Ensuring the validity of the feature space is crucial to obtain sufficiently similar instances. However, it may be difficult to achieve this when the feature representations are learned directly from MOS values. Specifically, the principle of visual masking effects~\cite{maskingneuronal,pattern,jndreview} highlights the influence of both image content and distortion on the perceptual function of the human visual system (HVS). Nevertheless, the MOS values primarily reflect the level of quality degradation without providing specific information about the content or distortion type. The image pairs exhibit varying image quality even if they have the same content (Fig. \ref{fig_dist_cont}~(a)) or the same distortion (Fig. \ref{fig_dist_cont}~(c)). Conversely, image pairs with different contents (Fig. \ref{fig_dist_cont}~(b)) or different distortions (Fig. \ref{fig_dist_cont}~(d)) may still have comparable image quality. These observations underline the inherent limitations of learning feature representations directly from MOS values. Therefore, BIQA models derived from regressing MOS values may not effectively model content and distortion representations, hindering their ability to mimic the perception function of the HVS.

	In this study, we introduce a regression-free BIQA model for image quality assessment. This model is built on the novel insight that images with similar content and distortion have similar perceptual quality. That is, the nearest neighbors in the feature space based on image content and distortion generally have close quality scores. The proposed model establishes a hierarchical $k$-NN algorithm to retrieve similar instances from a human-annotated IQA database. It contains two modules: the semantic classification (SC) module and the distortion classification (DC) module for retrieving image instances with similar semantic content and distortion. By aggregating subjective quality scores of retrieved instances, the model can predict image quality in a localized manner, avoiding the impact of non-representative training data on model parameters obtained through regression across all training samples.
	
	The main contributions of the paper can be summarized in three aspects as follows:
	\begin{itemize}
		\item We conduct an analysis of the quality relationship between image content and distortion and arrive at the conclusion that images with similar content demonstrate comparable image quality under the identical distortion. This discovery provides a valuable insight for enhancing the evaluation of image quality. 
		\item  We propose a regression-free method to assess image quality by retrieving similar instances in the feature space based on semantic and distortion features. The proposed approach establishes effective local neighbor relationships with high-level features obtained through classification, showing minimal performance degradation despite substantial feature dimensionality reduction.
		\item We evaluate the proposed model on seven benchmark IQA databases and the results demonstrate that our model outperforms regression-based BIQA models on learnable distortions and remains competitive in non-learnable distortion scenarios.
	\end{itemize}
The remainder of this paper is organized as follows: Section~\ref{relatedwork} provides a review of BIQA methods, including both regression-based and regression-free methods. Section~\ref{motivation} analyzes, with visual evidence, the relationship between image quality and the similarity of image content and distortion. The proposed regression-free BIQA framework is detailed in Section~\ref{method}. In Section~\ref{experiment}, we provide a comprehensive evaluation of current BIQA models, along with ablation studies and discussions. Finally, we provide a conclusion in Section~\ref{conclusion}.

\section{Related Work} \label{relatedwork}
In this section, we review previous work from two perspectives: regression-based BIQA approach and regression-free BIQA approach. Regression-based BIQA methods prioritize human perceptual data and aim to mimic human judgment through the training of parameterized models including traditional regression functions~\cite{biqi, diivine,bliinds-ii,brisque,cbiq, qac, qaf, lbiq, cornia, hosa} and deep neural networks~\cite{cnn, WaDIQaM, meon, rankiqa, deeprn, cahdc, sfa, hyperiqa, triq, tres, graphiqa, musiq, gao_csvt}, while regression-free BIQA methods rely on quality-centric features and computational analysis to assess image quality~\cite{niqe,ilniqe,cbiq,tclt,wacv_badu}. 
\subsection{Regression-based BIQA Methods}
In traditional BIQA methods, natural scene statistics (NSS)-based models~\cite{biqi, diivine,bliinds-ii,brisque,fang2015, wu2015} operate on the premise that natural images exhibit inherent statistical regularities that are disrupted by various distortions, which are quantified by assessing deviations from the statistical norm. Various hand-crafted features, such as discrete wavelet coefficients~\cite{biqi}, locally normalized luminance coefficients~\cite{brisque}, gradient magnitude and Laplacian of Gaussian response~\cite{gmlog}, have been proposed to identify distortions, which are then parameterized in distribution models, e.g., the generalized Gaussian distribution~\cite{biqi, brisque} and Weibull distribution~\cite{ilniqe}. These model parameters are subsequently employed in nonlinear regression functions such as SVR for image quality learning. In another approach, feature learning-based approaches~\cite{cbiq, qac, qaf, lbiq, cornia, hosa}, aim to autonomously derive quality-aware features from images, typically involving extracting local features, constructing a visual codebook to represent image, and performing quality regression. However, thees methods remain challenging to emulate the perceptual mechanisms of the HVS due to limitations in the representation of traditional features and the learning capabilities of traditional models. 

Deep learning-based BIQA models are becoming increasingly popular due to their powerful representation capabilities. Early methods were often trained directly on IQA datasets using random initialized  parameters~\cite{cnn,diqa,iqacnn+, biecon, WaDIQaM}. Kang et al.~\cite{cnn} introduced a shallow convolutional neural network (CNN) for quality learning using a patch-based approach. 
To address the challenge posed by the limited scale of IQA datasets and the risk of overfitting, data augmentation techniques have been proposed to expand the training data for pre-training BIQA models, including generating distorted images for classification~\cite{meon} and regression~\cite{cahdc, aigqa} or quality rankings~\cite{rankiqa, clriqa, listwiserank}. Then, the pre-trained models are fine-tuned by performing regression on human-annotated images with opinion scores to obtain the final BIQA models. Some approaches utilize deep semantic features for perceptual image quality modeling~\cite{deeprn, sfa, hyperiqa, triq, tres, graphiqa, musiq}. They employ pre-trained models from other visual tasks such as image classification~\cite{imagenet}, for feature extraction, paired with quality regression modules such as MLP~\cite{deeprn, sgdnet} hyper network~\cite{hyperiqa}, and Transformer encoder~\cite{tres,triq}.

\subsection{Regression-free BIQA Methods} 
Regression-free BIQA methods do not rely on human-rated scores for training~\cite{niqe,ilniqe,cbiq,tclt,wacv_badu}. These methods focus on intrinsic image properties, utilizing computational analysis to evaluate image quality. CBIQ~\cite{cbiq} employs Gabor filter features as local descriptors for patch clustering in training images and predicts quality for a test image by retrieving training images through matching descriptor distributions. TCLT~\cite{tclt} integrates statistics from various transform domains and color channels in images, using $k$-NN to query similar matches in human-annotated images for quality prediction. Babu et al.~\cite{wacv_badu} have proposed a content separation loss to reduce the content dependence in authentic image quality evaluation by quantifying the divergence between multivariate Gaussian parameters of distorted and undistorted images. Unlike previous approaches, we introduce feature spaces based on content and distortion, allowing the accurate prediction of quality by matching content and distortion consistency. Our method is built on the insight that images sharing similar content and distortion exhibit similar perceptual quality. The introduced feature spaces and similarity-based retrieve approach significantly outperform previous regression-free BIQA methods and even surpass regression-based methods.

\begin{figure}[!tpb]
	\centering
	\includegraphics[scale=.62]{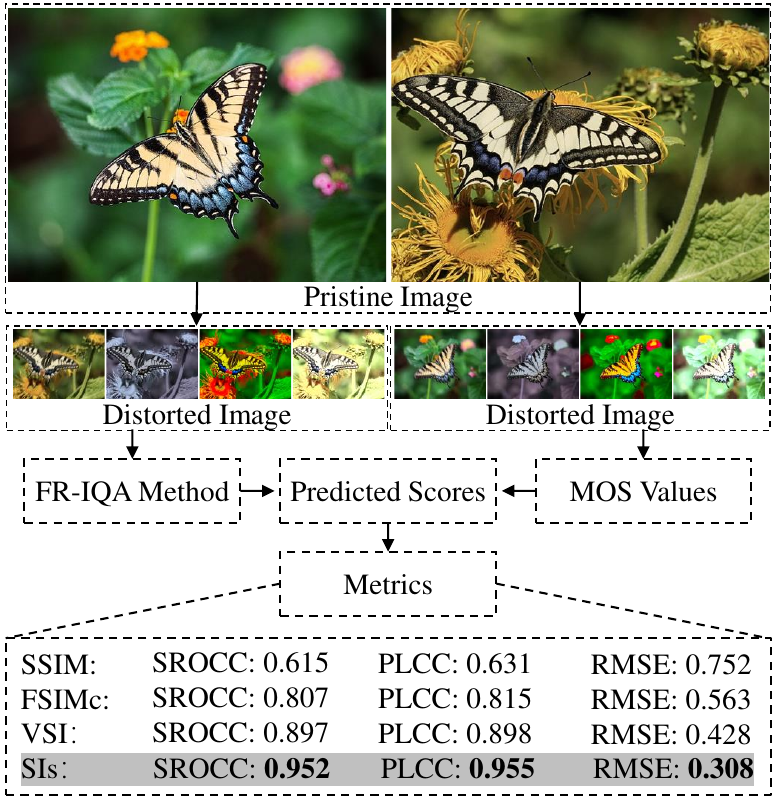} 
	\caption{Analyzing processes for quality correlation and precision between similar content images. The compared FR-IQA methods include SSIM, FSIMc, and VSI, with performance measured using SROCC, PLCC, and RMSE as evaluation metrics.} 
	\label{fig_sis}
\end{figure}	
\section{Analysis} \label{motivation} 
When perceiving identical images, the HVS will consistently produce the same visual responses.  Furthermore, images with similar content may produce similar perceptual properties in the HVS. That is, images with similar content degraded by the same distortion will have comparable visual quality. As illustrated in Fig.~\ref{fig_sis}, pristine image pair with similar content were obtained from KADID-10k~\cite{kadid-10k}. The distorted images corresponding to the left image in the image pair were used as input for several FR-IQA methods, including SSIM~\cite{ssim}, FSIMc~\cite{fsim}, and VSI~\cite{vsi}. The evaluation metrics SROCC, PLCC, and RMSE were calculated based on the predicted scores fitted by five-parameter logistic function~\cite{logistic_func} and mean opinion scores (MOS) values. The distorted images corresponding to the right image served as similar instances (SIs) for these input images, with the distortion types and levels aligned. The MOS values of the SIs were taken as the predicted quality scores. The results demonstrate that the SIs performed best in terms of correlation and precision, indicating that images with similar content generally have similar quality under identical distortions.

\begin{figure*}[htbp]
	\centering
	\subfigure[TID2013]{\begin{minipage}[t]{0.495\linewidth}
			\centering
			\includegraphics[width=\linewidth]{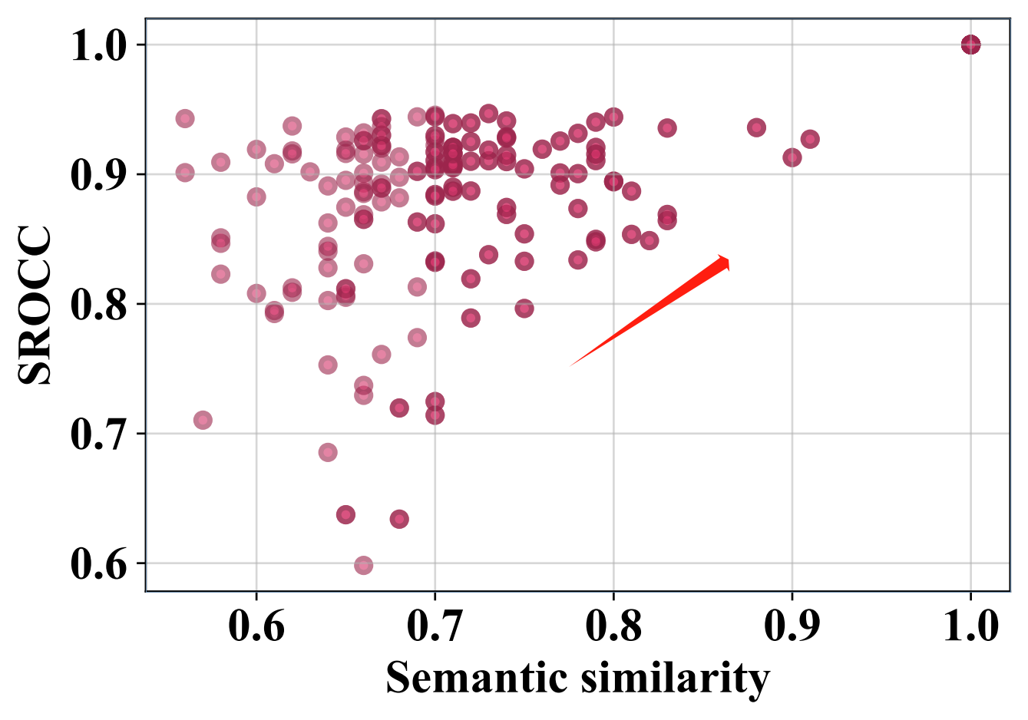}\noindent
		\end{minipage}}
	\subfigure[KADID-10k]{\begin{minipage}[t]{0.495\linewidth}
			\centering
			\includegraphics[width=\linewidth]{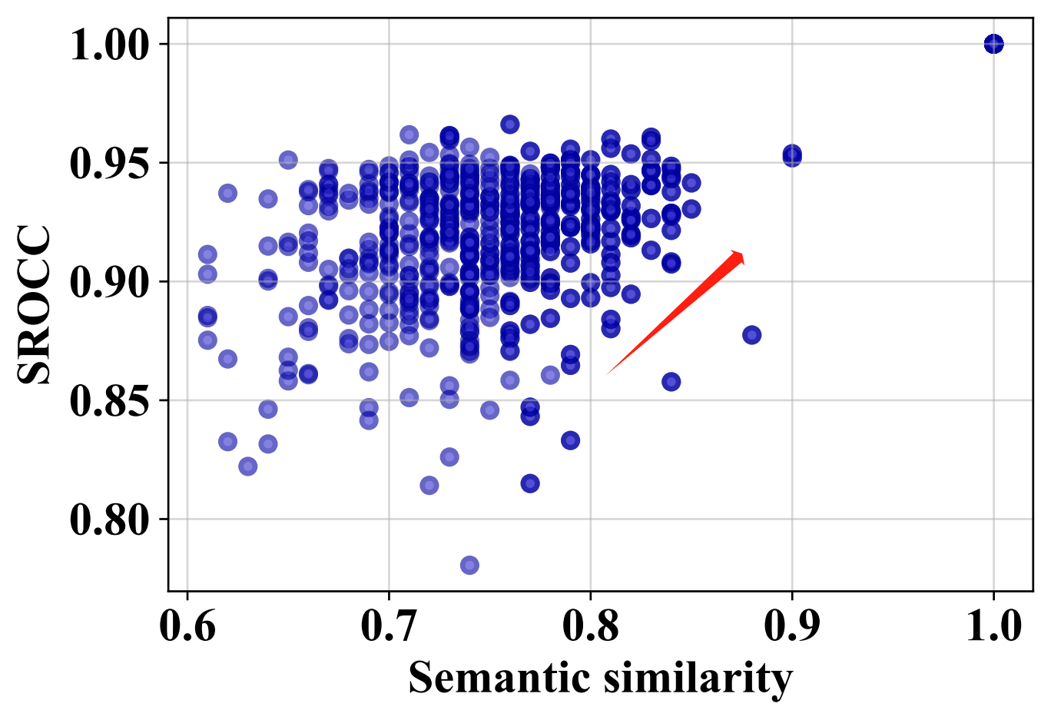}
		\end{minipage}}
		
%
	\centering
	\caption{Illustrations of the correlation between the quality of images with similar content and distortion. In these figures, the horizontal axis represents the semantic similarity between two pristine images. The vertical axis represents the quality relevance between the distorted versions of the two pristine images.}
	\label{fig_correlation}
\end{figure*}

To further investigate the quality correlation between images in both content and distortion, semantic feature vector were extracted by ResNet50~\cite{resnet} for each pristine image to represent its content. The content similarity between two pristine images is determined by the cosine distance between their semantic feature vectors. The correlation between the quality of distorted images associated with these two pristine images was measured using the SROCC of their MOS values, with aligned distortion types and levels. Fig.~\ref{fig_correlation} (a) and (b) illustrate the top  ten most semantically similar pristine images for each pristine image in the TID2013~\cite{tid2013} and KADID-10k~\cite{kadid-10k} datasets, respectively. In these figures, the horizontal axis represents the semantic similarity between two pristine images. The vertical axis represents the quality relevance between the distorted versions of the two pristine images. These figures demonstrate that the quality correlation, i.e., SROCC, increases with growing semantic similarity. Specifically, distorted images exhibit varying quality relevance given low semantic similarity. However, in regions with high semantic similarity (e.g., greater than 0.8), the quality relevance tends to be higher, indicating comparable perceptual quality for images of similar content and distortion. With this motivation, we introduce a regression-free BIQA method that evaluates image quality through the retrieval of content and distortion-similar instances. Further details will be presented in the next section.

\section{The Proposed Regression-free BIQA Method}\label{method}
\begin{figure*}[!tp]
	\centering
	\includegraphics[scale=.5]{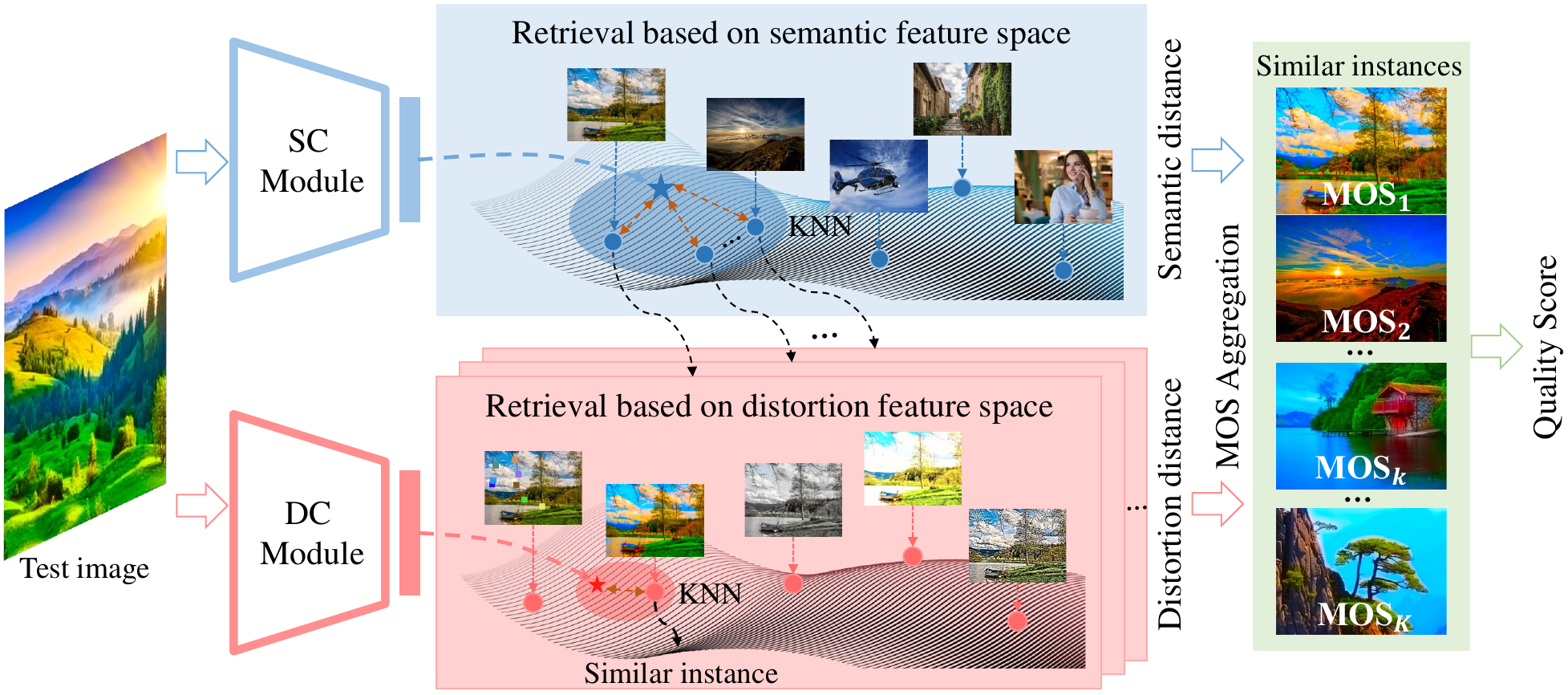} \vspace{-2mm}
	\caption{An overview of the proposed model, given an input image and an IQA database. The SC module is utilized to retrieve a pristine image with similar content. Then, the DC module is utilized to retrieve an instance with similar distortion within the distorted images associated with the retrieved pristine image. The quality prediction utilizes the MOS of the retrieved instance, and the final predicted score is computed by aggregating the MOS values from multiple instances retrieved using this approach.} 
	\label{fig_method}
	\vspace{-3mm}
\end{figure*} 
\subsection{Overview}
\noindent\textbf{Pipeline.} As illustrated in Fig.~\ref{fig_method}, we designed two classification modules, i.e., the semantic classification (SC) module and the distortion classification (DC) module, to identify the content and distortion of an image, respectively. Specifically, the SC module first employs a $k$-NN to retrieve pristine images that exhibit content similarity with the test image. Subsequently, the DC module employs a second $k$-NN to identify similar instances within the distorted images associated with each retrieved original image. Finally, the predicted quality score is obtained by aggregating the MOS values of these similar instances.

\noindent\textbf{Notations.}
Given an IQA database, following the sampling protocol for developing IQA models~\cite{bliinds-ii,cornia,sgdnet,hyperiqa}, the database is randomly divided into training and test sets, with an 80\%-20\% split based on pristine images. The training set, consisting of $N$ pristine images, is denoted as $\mathcal{D} =\left\{(x_i^{p}, \mathbf{D}_{i}, \mathbf{Y}_{i}) \right\}_{i=1}^N $, where $x_i^{p}$ denotes the $i$-th pristine image, $\mathbf{D}_i= \left\{x_{ij}\right\}_{j =1}^M$ denotes the distorted images corresponding to $x_i^{p}$, with $x_{ij}$ denoting the $j$-th sample in $\mathbf{D}_i$, and $\mathbf{Y}_i = \left\{y_{ij}\right\}_{j=1}^M$ denotes the MOS values corresponding to these distorted images.
\subsection{Semantic Classification (SC) Module}
In this subsection, we describe the design of the SC module, which is tasked with extracting the semantic vector to represent image content. Limited the number of pristine images in IQA databases, e.g., TID2013~\cite{tid2013} with only 25 pristine images, the semantic vector distance is used to represent the similarity of image content instead of finding an identical image category. Specifically, the ResNet-18~\cite{resnet} network pre-trained on ImageNet~\cite{imagenet} without the last classification layer is taken as the SC module for extracting semantic vectors. Let $\mathcal{E}_s$ represent the SC module with parameter $\omega_s$. Given a test image $x$ and a pristine image $x_i^p$, the content similarity between $x$ and $x_i^p$ is measured as the semantic distance $\mathbf{d}_s(x, x_i^p)$, which is defined as the cosine distance of their semantic feature vectors by
\begin{equation}\label{eqn:eq2}
\mathbf{d}_s(x_t, x_i^p) =  1 - \frac{\mathcal{E}_s(x; \omega_s) \cdot \mathcal{E}_s(x_i^p; \omega_s)}{|\mathcal{E}_s(x; \omega_s)| |\mathcal{E}_s(x_i^p; \omega_s)|}
\end{equation}
Then, the top $k'$ pristine images that exhibit the highest similarity to $x$ are retrieved by
\begin{equation}
	\left\{\hat{x}_i^{p}\right\}_{i=1}^{k'} = \left\{\hat{x}_i^p| 	\mathop{\mathrm{min_\mathbb{K}}}\limits_{ \hat{x}_i^p \in \left\{x_i^{p}\right\}_{i=1}^N} \mathbf{d}_s(x, \hat{x}_i^{p})\right\},
\end{equation}
where $1 \leq k' \leq N$. The function $\mathrm{min_\mathbb{K}}(\cdot)$ selects samples associated with the $k$ smallest values. Note that when evaluating the model on the IQA dataset, the original image corresponding to $x$ is not contained in the set $\left\{x_i^{p}\right\}_{i=1}^N$.

\subsection{Distortion Classification (DC) Module}
In this subsection, we describe the design of the DC module, which is obtained by training a distortion classification model~\cite{boatnet} on the IQA datasets. The model was trained on two distortion scenarios: single distortions containing one distortion type, and mixed distortions with multiple distortion types. Single distortions enable analyzing the impact of individual distortion types on perceptual quality, while mixed distortions are designed for evaluating model performance in real-world scenarios.

\noindent\textbf{Single distortion.}~
Given a training set denoted as $\left\lbrace\left(x^n, \mathbf{p}^n\right)| 1 \leq n \leq N\times M\right\rbrace$, consisting of $N$ pristine reference images, each with $M$ corresponding distorted versions. This yields a total of $N\times M$ training samples, where $x^n$ represents the $n$-th sample and $\mathbf{p}^n$ represents the ground-truth distortion class indicator vector. The training objective is to optimize the model parameters $\omega_d$ by minimizing the cross-entropy loss function, which is defined as
\begin{equation}\label{eqn:eq3}
	\ell_{sd} = -\sum_{n}\sum_{c}\mathbf{p}^n_c \cdot \log(\mathbf{\hat{p}}(x_c^n;~\omega_d))
\end{equation}
where $c$ indexes the distortion categories.

\noindent\textbf{Mixed distortion.} We used two-stage degradation\cite{practical_deg} to synthesize the mixed distortion image. In each of the two degradation stages, we randomly apply 1 to 3 distortions, e.g., blurring, brighten, darken, and JPEG compression, setting the corresponding indices in the multi-hot encoded label vectors to 1 for the chosen distortions. Detailed distortion codes are available in the published code. Pre-training was performed on 100,000 images from KADIS-700k~\cite{kadid-10k}, with the training objective being optimization of model parameters $\omega_d$ by minimizing a multi-label classification loss, which is defined as
\begin{equation}\label{eq: loss_md}
	\ell_{md} = -\sum_{n}\sum_{c}\mathbf{p}^n_c \cdot \log(\mathbf{\hat{p}}(x_c^n;~\omega_d))+(1- \mathbf{p}^n_c) \cdot \log(1- \mathbf{\hat{p}}(x_c^n;~\omega_d))
\end{equation}

Then, the pre-trained model is fine-tuned on authentic distortion data. The labels of distortion types are generated by transforming the predicted distortion probability vector $\mathbf{\hat{p}}^n_c$ into binary labels using the function $\mathbf{1}[\cdot]$, which assigns a value of 1 when the probability exceeds the threshold $\tau$.
\begin{equation}\label{eq8}
	\mathbf{p}^n_c = \mathbf{1}[\mathbf{\hat{p}}^n_c > \tau],
\end{equation}

 Eq.~(\ref{eq: loss_md}) is only employed for the classification of distortion types. To evaluate the degree of distortions, we discretize continuous MOS scores into different distortion levels. The ground-truth level class indicator vector is represented as $\mathbf{L}^n_l$. The loss function for level classification is defined as:
\begin{equation}\label{eq9}
	\ell_{l} = -\sum_{n}\sum_{l}\mathbf{L}^n_l \cdot \log(\mathbf{\hat{L}}(x_l^n;~\omega_d)),
\end{equation}
where the value of $l$ ranges from 1 to 10. Finally, the pre-trained weights are fine-tuned through a combined loss function of $\ell_{md} + 2 \ell_{l}$.

 The final classification layer of the trained model is removed, and the resulting model is used as the DC module to extract the distortion features. The distortion similarity between the test image $x$ and a distorted image $x_{ij}$ is calculated as the cosine distance between their distortion feature vectors by
\begin{equation}\label{eqn:eq2}
	\mathbf{d}_d(x, x_{ij}) =  1 - \frac{\mathcal{E}_d(x; \omega_d) \cdot \mathcal{E}_d(x_{ij}; \omega_d)}{|\mathcal{E}_d(x; \omega_d)| |\mathcal{E}_d(x_{ij}; \omega_d)|}
\end{equation}
Given a pristine image $\hat{x}_i^p$ retrieved from the SC module, its associated set of distorted images $\mathbf{\hat{D}}_{i}$ can be acquired. Among these distorted images, the one with the highest distortion similarity to $x$ is selected by
\begin{equation}\label{eq11}
	\left\{\hat{x}_{ij}\right\}_{j=1}^{k''}= \left\{\hat{x}_{ij}| 	\mathop{\mathrm{min_\mathbb{K}}}\limits_{ \hat{x}_{ij} \in \mathbf{\hat{D}}_{i}} \mathbf{d}_d(x, \hat{x}_{ij})\right\},
\end{equation}
Where the $k''$ is set to 1 by default. The distorted image $\hat{x}_{ij}$ is taken as the instance that has a similar perception to~$x$ in both content and distortion. Multiple similar instances can be obtained based on different retrieved pristine images. Notably, since authentic distortion images lack pristine reference, we directly retrieve similar instances from training set by concatenating semantic and distortion features.

\subsection{Image Quality Prediction}
Aggregating the MOS values of multiple retrieved instances can mitigate prediction bias, as a single instance may not be representative of the image being evaluated in all cases. We propose two alternative quality aggregation strategies,~i.e., simple average and weighted average, to evaluate the image quality score.

\noindent\textbf{Simple average.}
The prediction score for $x$ is computed by taking the average of the MOS values associated with the retrieved instances, which is given by
\begin{equation}\label{eq12}
\mathcal{Q}_{s} = \dfrac{1}{k' k''}\sum_{i,j} \text{y}_{ij}, \\
\end{equation}
where $\text{y}_{ij}$ represents the ground truth MOS value of the instance $x_{ij}$. $k''$ is set to 1 by default, resulting in a total of $k'\cdot k'' = k'$ instances.

\noindent\textbf{Weighted average.}~The weighted average approach considers the relative impact of each retrieved instance on the final quality prediction. The inverse of the sum of semantic distance and distortion distance between the test instance $x$ and each retrieved instance $x_{ij}$ were used to calculate weighting coefficients for averaging their MOS values $y_{ij}$. This assigns greater weights to instances more proximate to $x$ in semantic and distortion feature space. Finally, the predicted quality score can be expressed as
\begin{equation}\label{eq13}
	\mathcal{Q}_{w} = \frac{\sum_{i, j} [\mathbf{d}_s(x, \hat{x}_i^{p}) + \mathbf{d}_d(x, \hat{x}_{ij})]^{-1} \cdot \text{y}_{ij}}{\sum_{i, j} [\mathbf{d}_s(x, \hat{x}_i^{p}) + \mathbf{d}_d(x, \hat{x}_{ij})]^{-1}} \\
\end{equation}
\section{Experimental Results}\label{experiment}
\subsection{Databases}
We conducted a comprehensive performance evaluation of the proposed model using seven IQA benchmark databases including three different image types: synthetically distorted images, authentically distorted images, and algorithmically processed images. The evaluation utilized four synthetic distortion databases, namely CSIQ~\cite{csiq}, TID2013~\cite{tid2013}, KADID-10k~\cite{kadid-10k} and LIVE-MD~\cite{livemd}. The images in LIVE-MD are degraded by multiple distortion types, while the other synthetic databases consist of images degraded by a single distortion category. In addition, two authentic distortion databases were tested, namely CLIVE~\cite{livec} and KonIQ-10k~\cite{koniq-10k}. Furthermore, the recent PIPAL database~\cite{pipal} containing algorithmically processed images was also included in our benchmarking. Table~\ref{tab:tab1} summarizes the key characteristics of these employed IQA databases, including the number of reference images, distortion types, distorted images, and subjective quality labels.
\subsection{Evaluation criteria}
The performance of BIQA models is commonly evaluated using two correlation criteria, namely Spearman's rank order correlation coefficient (SROCC) and Pearson's linear correlation coefficient (PLCC). The SROCC measures the monotonic relationship between the ground truth and predicted scores, represented as:
\begin{equation}
	{\rm SROCC} = 1 - \frac{6 \sum ^K_{i=1} d_i^2}{K(K^2 - 1)},
\end{equation}
where $K$ is the number of testing images, and $d_i$ is the rank difference between the predicted score and the ground truth value. The PLCC, on the other hand, measures the linear correlation between the ground truth and predicted scores, expressed as:
\begin{equation}
	{\rm PLCC} =\frac{\sum ^K_{i=1}(q_i - u_{q})(\hat{q}_i - u_{\hat{q}})}{\sqrt{\sum ^K_{i=1}(q_i - u_{q})^2} \sqrt{\sum ^K_{i=1}(\hat{q}_i - u_{\hat{q}})^2}},
\end{equation}
where $q_i$ and $\hat{q}_i$ denote the ground truth and the predicted score of the $i$-th image, respectively, and $u_{q}$ and $u_{\hat{q}}$ are the corresponding mean values, represented as $u_{q}=\sum_{i=1}^{K}q_i$ and $u_{\hat{q}}=\sum_{i=1}^{K}\hat{q}_i$, respectively.
\subsection{Implementation details}
\begin{table}[!tbp]
	\centering
	\caption{The benchmark databases for BIQA model evaluation.}
	\fontsize{10.pt}{10.pt}\selectfont%
	\renewcommand{\arraystretch}{1.05}
	\resizebox{3.4 in}{!} 	{
		\begin{tabular}{ccccc}
			\toprule
			\multirow{2}{*}{\textbf{Database}} & \textbf{\# of Ref}. & \textbf{\# of Dist}. & \textbf{\# of Dist}. & \textbf{Score}\tabularnewline
			& \textbf{Images} & \textbf{Types} & \textbf{Images} & \textbf{Type}\tabularnewline
			\midrule
			CSIQ \cite{csiq} 	 & 30 & 6 & 866 & DMOS\tabularnewline
			TID2013 \cite{tid2013} & 25 & 24 & 3,000 & MOS\tabularnewline
			KADID-10k \cite{kadid-10k} & 81 & 25 & 10,125 & MOS\tabularnewline
			LIVE-MD \cite{livemd}& 15 & 2 & 150 & DMOS\tabularnewline
			\midrule
			CLIVE \cite{livec}& - & - & 1162 & MOS\tabularnewline
			KonIQ-10k \cite{koniq-10k}& - & - & 11073 & MOS\tabularnewline
			\midrule
			PIPAL~\cite{pipal} & 250 & 40 & 29k & MOS\tabularnewline
			\bottomrule
		\end{tabular}	
	}
	\label{tab:tab1}
\end{table}

\begin{table*}[htp!]
	\centering
	\caption{Comparison with state-of-the-art BIQA models on synthetic distortion and authentic distortion databases, where bold \textcolor{VioletRed}{\textbf{pink}} and \textbf{black} entries indicate best and second best performance.}
	\fontsize{9.pt}{9.pt}\selectfont%
	\renewcommand{\arraystretch}{1}
	\resizebox{7. in}{!}{
		\begin{tabular}{ccccccccccccc}
			\toprule
			\multirow{2}{*}{\textbf{Method}} & \multicolumn{2}{c}{\textbf{CSIQ}\cite{csiq}} & \multicolumn{2}{c}{\textbf{TID2013}\cite{tid2013}}& \multicolumn{2}{c}{\textbf{KADID-10k}\cite{kadid-10k}}  & \multicolumn{2}{c}{\textbf{LIVEMD}\cite{livemd}}&\multicolumn{2}{c}{\textbf{CLIVE}~\cite{livec}} & \multicolumn{2}{c}{\textbf{KonIQ-10k}\cite{koniq-10k}}\\
			\cmidrule(r){2-3}\cmidrule(r){4-5}\cmidrule(r){6-7}\cmidrule(r){8-9} \cmidrule(r){10-11} \cmidrule(r){12-13}
			\multirow{2}{*}{} & \textbf{SROCC$\uparrow$} & \textbf{PLCC$\uparrow$} & \textbf{SROCC$\uparrow$} & \textbf{PLCC$\uparrow$} & \textbf{SROCC$\uparrow$} & \textbf{PLCC$\uparrow$} & \textbf{SROCC$\uparrow$} & \textbf{PLCC$\uparrow$} & \textbf{SROCC$\uparrow$} & \textbf{PLCC$\uparrow$}&\textbf{SROCC$\uparrow$} & \textbf{PLCC$\uparrow$} \tabularnewline
			\midrule
			DIIVINE~\cite{diivine} & 0.784& 0.836 &0.654& 0.549& 0.532& 0.489&0.874&0.894& 0.588  &0.591&0.546& 0.558
			\tabularnewline
			BRISQUE\cite{brisque} & 0.750 & 0.829 & 0.604 & 0.651 & 0.554 &0.519 &0.897 &0.921&0.629  &0.629&0.681& 0.685
			\tabularnewline
			HOSA~\cite{hosa}&0.781 &0.842& 0.688 &0.764 &0.653& 0.609&0.913&0.926&0.621& 0.657& 0.684& 0.703 \tabularnewline
				\midrule
			ILNIQE~\cite{ilniqe}& 0.806 &0.808& 0.521& 0.648& 0.558 &0.534 &0.902&0.914&0.432& 0.508 &0.507 &0.523\tabularnewline
			Babu et. al.~\cite{wacv_badu}&-&-&-&-&-&-&-&-&0.51& 0.52 &0.65& 0.64\tabularnewline
				\midrule
			MEON~\cite{meon}&0.852&0.864&0.808&0.824&0.604&0.691&0.924&0.940&0.613 &0.613 & 0.651&0.654 \tabularnewline
			MetaIQA\cite{meta}&0.899& 0.908& 0.856 & 0.868&0.762&0.775&-&-&0.802 &0.835& 0.856 &0.887\tabularnewline
			CLRIQA\cite{clriqa}&0.915& 0.938& 0.837& 0.863& 0.837& 0.843&-&-&0.832 &0.866 &0.831 &0.846\tabularnewline
			Gao et.al.\cite{gao_csvt}&0.933& 0.948&0.844 &0.852&-&-& 0.925& 0.938&0.830& 0.863&\textbf{0.908}& \textbf{0.921}\tabularnewline
			CaHDC\cite{cahdc}& 0.903 & 0.914 & 0.862 & 0.878 & - & - & 0.927 & \textcolor{VioletRed}{\textbf{0.950}} &0.738 &0.744 & - & - \tabularnewline
			AIGQA\cite{aigqa}  & 0.927 & 0.952 &\textbf{0.871} & \textbf{0.893} & \textbf{0.864} & \textbf{0.863} & 0.933 & \textbf{0.947}&0.751&0.761& -&- \tabularnewline
			DB-CNN\cite{dbcnn}& \textbf{0.946} & \textcolor{VioletRed}{\textbf{0.959}}&0.816 &0.865 &0.851 &0.856&0.927&0.934 &\textbf{0.851}& 0.869& 0.875 &0.884	\tabularnewline
			HyperIQA\cite{hyperiqa}&0.923 &0.942&0.840& 0.858&0.852  &0.845 &\textcolor{VioletRed}{\textbf{0.939}}&0.938&\textcolor{VioletRed}{\textbf{0.859}}&\textcolor{VioletRed}{\textbf{0.882}}& 0.906 &0.917\tabularnewline
			TReS\cite{tres}& 0.922& 0.942&0.863&0.883&0.859 &0.858&0.930&0.932& 0.846&\textbf{0.877}& \textcolor{VioletRed}{\textbf{0.915}}& \textcolor{VioletRed}{\textbf{0.928}} \tabularnewline
			\midrule
			Our&\textcolor{VioletRed}{\textbf{0.956}}&\textbf{0.954}&\textcolor{VioletRed}{\textbf{0.918}}&\textcolor{VioletRed}{\textbf{0.921}}&\textcolor{VioletRed}{\textbf{0.947}}&\textcolor{VioletRed}{\textbf{0.949}}&\textbf{0.936} & \textbf{0.947}&0.758&0.780&0.868&0.885 \tabularnewline
			\bottomrule
		\end{tabular}
	}
	\label{single_dataset}
\end{table*}

We leveraged PyTorch \cite{pytorch} as the codebase to implement the proposed model. We conducted the experiments using the NVIDIA RTX 3090 GPU. For mixed-distortion classification models, we trained on a subset of 100,000 images from KADIS-700k~\cite{kadid-10k}, with 5,000 used for validation. Training employed an SGD optimizer with a weight decay of 1e-4 and momentum of 0.9. The initial learning rate was set at 0.05 and adjusted using a cosine learning rate scheduler. The training spanned 50 epochs, with a batch size of 32. Training samples were randomly cropped to a size of 288 $\times$ 384 pixels from the input images. For single-distortion classification models, we utilize pre-trained weights from an extensive dataset in prior work~\cite{vipnet} for model initialization. The pre-trained distortion classification model was fine-tuned on IQA datasets for 30 epochs using the SGD optimizer, with a mini-batch size of 16 and an initial learning rate of 5e-3. The learning rate was reduced by a factor of 0.5 after every eight epochs. The trained distortion classification model without the final linear layer is adopted as the DC module. The ResNet-18 pre-trained in ImageNet without the classification layer is adopted as the SC module. Due to the limited size of the LIVEMD dataset, we set $k'$ to 4, 10 for other synthetic distortion datasets and PIPAL dataset, and 15 for authentic distortion datasets. The threshold $\tau$ in Eq.~\ref{eq8} defaults to 0.2. In the image quality evaluation process, the test image was first fed directly to the SC module, and then randomly cropped to a size of 288~$\times$~384 before being input to the DC module. To avoid content overlap in the synthetic distortion dataset, two subsets of pristine images were randomly divided in an 8:2 ratio, and corresponding distorted images were allocated accordingly to form training and testing sets. For authentic distortion datasets, the entire dataset was divided into two subsets with an 8:2 ratio. To avoid performance bias, the experiments were conducted 15 times, with random operations, and the median of SROCC and PLCC results were reported. Notably, the training set was only used for training the DC module and retrieving instances, not for quality regression.
\begin{table}[tp!]
	\centering
	\caption{Performance comparison of BIQA methods on PIPAL \cite{pipal}.}
	\fontsize{8.5pt}{8.5pt}\selectfont%
	\renewcommand{\arraystretch}{1.05}
	\resizebox{3.4 in}{!} 	{
		\begin{tabular}{cccccc}
			\toprule
			\multicolumn{3}{c}{\textbf{PIPAL Training Set (8:2 ratio)}} & \multicolumn{3}{c}{\textbf{PIPAL-NTIRE22-Test}\cite{NTIRE2022}} \\
			\cmidrule(r){1-3}\cmidrule(r){4-6}
			\textbf{Method} & \textbf{SROCC$\uparrow$} & \textbf{PLCC$\uparrow$}& \textbf{Rank}&\textbf{SROCC$\uparrow$} & \textbf{PLCC$\uparrow$} \tabularnewline
			\midrule
			GraphIQA~\cite{graphiqa} &0.483&0.522&1&0.7040 &0.7396   \tabularnewline
			DB-CNN~\cite{dbcnn}&0.577&0.576&2& 0.6996 &0.7371 \tabularnewline
			HyperIQA~\cite{hyperiqa}&0.593 &0.639&3& 0.6965 &0.7254 \tabularnewline
			TReS~\cite{tres}&0.601 &0.655 &4&  0.6808 &0.7257 \tabularnewline
			\midrule
			Ours&\textbf{0.7279} &\textbf{0.7597} &-& -&- \tabularnewline
			\bottomrule
	\end{tabular}	}
	\label{pipal}
\end{table}

\subsection{Comparison with the State-of-the-art Methods}
To assess the performance of the proposed model, experiments were carried out on seven IQA databases and the results were compared with 12 regression-based BIQA methods, including three traditional methods~\cite{diivine,brisque, hosa} and nine deep learning-based methods~\cite{meon,meta,clriqa,gao_csvt,cahdc,aigqa, dbcnn,hyperiqa,tres}, and two regression-free methods~\cite{ilniqe, wacv_badu}. For the traditional BIQA methods, the results are reproduced by the source code released by the authors. For the deep learning-based methods, the results of the corresponding database are sourced from the original publication or reference~\cite{tres}, while the results of the remaining database are reproduced from the published source code.

\noindent\textbf{Evaluation on synthetic distortion databases.}~
As shown in Table~\ref{single_dataset}, the proposed model demonstrates superior performance across four synthetic distortion datasets, either achieving state-of-the-art results or approaching the best methods. For the relatively small CSIQ and LIVE-MD datasets, our model is on par with the top-performing BIQA methods. Specifically, when testing on CSIQ, our method achieves the best SROCC of 0.956, exceeding DB-CNN by 1.1\%, while DB-CNN~\cite{dbcnn} achieves the best PLCC, surpassing us by 0.5\%. On LIVE-MD, HyperIQA~\cite{hyperiqa} and CaHDC~\cite{cahdc} obtained the highest SROCC of 0.939 and PLCC of 0.950 respectively, with our approach achieving highly comparable SROCC and PLCC results of 0.936 and 0.947 respectively. However, on the relatively larger TID2013 and KADID-10k datasets, our method significantly outperforms existing BIQA methods in both SROCC and PLCC. Specifically, on TID2013 our approach surpasses the second best method AIGQA~\cite{aigqa} by margins of 5.4\% SROCC and 3.1\% PLCC. Similarly, on KADID-10k, we exceed the second best method by even wider gaps, achieving a 9.6\% higher SROCC and a 10.0\% higher PLCC.

This superior performance on larger datasets can be attributed to two main factors. First, the increased dataset size offers a broader range of data points, which improves the accuracy and reliability of our similar instance-based prediction. Second, as the dataset size increases, the regression models require optimization of the overall performance across all training samples. Thus, they must reconcile larger amounts of individual sample prediction errors, which leads to increased uncertainty and instability in predictions on specific samples. In contrast, our approach always retrieves the most similar instances of each sample individually, allowing for more consistent quality predictions and showing its advantages at scale.

\noindent\textbf{Evaluation on authentic distortion databases.}~Our approach demonstrates significant improvements over previous regression-free methods~\cite{ilniqe, wacv_badu}. Specifically, compared to Babu et al.'s method~\cite{wacv_badu}, our approach achieves a 48.6\% increase in SROCC and a 50.0\% increase in PLCC on CLIVE. Similarly, we observe a 33.5\% improvement in SROCC and a 38.3\% improvement in PLCC on KonIQ-10k. While our performance slightly trails behind some state-of-the-art methods such as  HyperIQA~\cite{hyperiqa} and TReS\cite{tres}, it surpasses several deep learning regression-based BIQA models. For instance, on the CLIVE dataset, our method outperforms MEON~\cite{meon}, CaHDC~\cite{cahdc}, and AIGQA~\cite{aigqa}, and on the KonIQ-10k dataset, it outperforms MEON~\cite{meon} and CLRIQA~\cite{clriqa} and performs comparably with MetaIQA~\cite{meta} and DB-CNN~\cite{dbcnn}. Notably, our model has not been trained using images with subjective quality scores.

\begin{table*}[!htbp] \centering
	\caption{SROCC evaluations on the individual distortions of TID2013. The number of times (N.T) each method achieves the best performance is listed in the last column.}
	\scalebox{0.95}[0.95]{
		\renewcommand{\arraystretch}{1.}
		\begin{tabular}{cccccccccccccc}
			\toprule
			\textbf{Method} &\textbf{AGN}& \textbf{ANC} &\textbf{SCN}&\textbf{MN}&\textbf{HFN}&\textbf{IN}&\textbf{QN}&\textbf{GB}& \textbf{DEN}&\textbf{JPEG}&\textbf{JP2K}&\textbf{JGTE}&\textbf{J2TE} \tabularnewline
			\midrule
			DB-CNN\cite{dbcnn} &0.790 &0.700 &0.826 &0.646 &0.879 &0.708 &0.825 &0.859 &0.865 &0.894 &0.916&0.772&0.773 \tabularnewline
			HyperIQA \cite{hyperiqa} &0.803 &0.503 &\textbf{0.980} &0.231 &0.857 &0.775 &0.877 &0.777 &0.843 &0.863 &0.900 &0.789 &0.843  \tabularnewline
			TReS \cite{tres} &0.911 &0.858 &0.938 &0.620 &0.875 &0.701 &0.878 &0.880 &0.883 &0.840 &\textbf{0.941} &0.753 &\textbf{0.884}  \tabularnewline
			CLRIQA \cite{clriqa}& 0.816 &0.720& 0.860 &0.395 &0.912&0.898&0.879& 0.892 &0.866& 0.940 &0.894 &0.785 &0.865 \\
			\midrule
			Ours &\textbf{0.946}& \textbf{0.872}& 0.956& \textbf{0.838}&\textbf{0.938}& \textbf{0.913} &\textbf{0.954}& \textbf{0.928}&\textbf{0.928}& \textbf{0.943} &0.934 &\textbf{0.845}& 0.832 \tabularnewline
			\toprule
			\textbf{Methods} &\textbf{NEPN}&\textbf{Block}&\textbf{MS}&\textbf{CTC}&\textbf{CCS}&\textbf{MGN}&\textbf{CN}&\textbf{LCNI}&\textbf{ICQD} & \textbf{CHA}&\textbf{SSR}&\multicolumn{2}{|c}{\textbf{N.T}} \tabularnewline
			\midrule
			DB-CNN\cite{dbcnn} &0.270 &0.444 &-0.009 &0.548 &0.631&0.711 &0.752 &0.860 &0.833 &0.732 &0.902 &\multicolumn{2}{|c}{0} \tabularnewline
			HyperIQA \cite{hyperiqa} &0.295 &0.155 &0.228 &0.705 &0.559 &0.874 &0.815 &0.902 &0.820 &0.865 &0.923 &\multicolumn{2}{|c}{1}\tabularnewline
			TReS \cite{tres} &0.358 &0.688 &0.344 &0.753 &\textbf{0.735}&0.815 &0.729 &0.890 &0.805 &0.798 &\textbf{0.947}  &\multicolumn{2}{|c}{4} \tabularnewline
			CLRIQA \cite{clriqa} &0.764&\textbf{0.697} &0.387& 0.784 &0.710& 0.801 &0.849 &0.901 &0.894 &\textbf{0.903} &0.919&\multicolumn{2}{|c}{2}  \\
			\midrule
			Ours &\textbf{0.797}& 0.596 &\textbf{0.688}& \textbf{0.847} &0.662 &\textbf{0.927} &\textbf{0.851} &\textbf{0.960}& \textbf{0.906}& 0.844 &0.928&\multicolumn{2}{|c}{\textbf{17}}\tabularnewline
			\bottomrule
		\end{tabular}
	}
	\label{dist_type_tid2013}
\end{table*}
\begin{table*}[htbp]
	\centering
	\caption{SROCC evaluations on the individual distortions of Kadid-10k. The number of times (N.T) each method achieves the best performance is listed in the last column.} 
	\scalebox{0.9}[0.9]{
		\renewcommand{\arraystretch}{1.1}
		\begin{tabular}{cccccccccccccc}
			\toprule
			\textbf{Method} &\textbf{GB}&\textbf{LB}&\textbf{MB}&\textbf{CD}&\textbf{CS}&\textbf{ICQD}&\textbf{CCS}&\textbf{CCS-Lab}&\textbf{JP2K}&\textbf{JPEG}&\textbf{AGN}&\textbf{ANC}&\textbf{IN}\tabularnewline
			\midrule
			DB-CNN \cite{dbcnn}  &0.838 &0.787 &0.645 &0.785 &0.622 &0.796 &0.170 &0.818 &0.748 &0.742 &0.817 &0.853 &0.706 \tabularnewline
			HyperIQA \cite{hyperiqa} &0.902&0.925&0.894&0.905&0.832&0.739&0.511&0.877&0.846&0.819&0.765&0.810&0.833  \tabularnewline
			TReS \cite{tres} &0.923&\textbf{0.938}&0.926&\textbf{0.908}&\textbf{0.889}&0.747&0.543&0.858&0.867&0.863&0.792&0.836&0.846  \tabularnewline
			\midrule
			Ours&\textbf{0.954}& 0.928& \textbf{0.952}& 0.828&0.789& \textbf{0.820} &\textbf{0.555}& \textbf{0.910}&
			\textbf{0.923} &\textbf{0.906} &\textbf{0.894}& \textbf{0.929}& \textbf{0.900} \tabularnewline
			\toprule
			\textbf{Method} &\textbf{MGN}&\textbf{DEN}&\textbf{Brighten}&\textbf{Darken}&\textbf{MS}&\textbf{Jitter}&\textbf{NEPN}&\textbf{Pixelate}&\textbf{QN}&\textbf{Block}&\textbf{HS}&\textbf{CTC}&\multicolumn{1}{|c}{\textbf{N.T}} \tabularnewline
			\midrule
			DB-CNN \cite{dbcnn} &0.753 &0.832 &0.629 &0.493 &0.415 &0.489 &0.309 &0.627 &\textbf{0.748} &0.361 &0.469 &0.388 &\multicolumn{1}{|c}{1} \tabularnewline
			HyperIQA \cite{hyperiqa} &0.860&\textbf{0.928}&0.811&0.622&0.317&0.882&0.473&0.803&0.723&0.368&0.919&0.333&\multicolumn{1}{|c}{1} \tabularnewline
			TReS \cite{tres}        &0.883&0.912&0.805&0.628&0.396&0.896&0.385&0.692&0.674&\textbf{0.574}&0.879&0.408&\multicolumn{1}{|c}{4}  \tabularnewline
			\midrule
			Ours &\textbf{0.907} &0.922& \textbf{0.937}& \textbf{0.917} &\textbf{0.655}& \textbf{0.945}&\textbf{ 0.625}& \textbf{0.831}&0.728& 0.531 &\textbf{0.933}& \textbf{0.796}&\multicolumn{1}{|c}{\textbf{19}} \tabularnewline
			\bottomrule
	\end{tabular}}
	\label{dist_type_kadid}
\end{table*}
The reasons for our approach falling slightly short of the best methods are twofold. First, real-world distorted images often involve unknown and complex distortion types, posing a challenge for the DC module in effectively modeling these distortions. Second, authentic distortion datasets have a wide range of image content and distortions, making it more challenging to retrieve highly similar instances. This challenge can typically be mitigated by increasing the number of instances to be retrieved. For instance, KonIQ-10k dataset is larger than CLIVE dataset and has superior performance compared to the CLIVE dataset. Nonetheless, compared to regression-based models, our proposed method achieves decent performance on authentic distortion datasets, without the need for training on images with quality scores.

\noindent\textbf{Evaluation on algorithmic generation database.}~One of the most important applications of IQA is measuring the performance of image restoration algorithms. The PIPAL dataset introduces a new distortion type based on GAN-generated outputs. As publicly available labels were unavailable for the PIPAL test set, we divided the training data into an 80:20 ratio based on the pristine images for model development and testing. Our approach is benchmarked against four existing BIQA methods~\cite{graphiqa, dbcnn, hyperiqa, tres} by reproducing results using released code in the same dataset splits. As shown in Table~\ref{pipal}, our proposed method surpasses the second-best performer, TReS, by margins of 21.1\% and 16.0\% for SROCC and PLCC respectively. In addition, top-four challenge rankings from the 2022 PIPAL NTIRE \cite{NTIRE2022} competition are presented. While our proposed method was not assessed on the PIPAL test set consisting of 1650 images, it holds promise. The reason is that we only used a subset of 80\% of the training data, with the remaining 20\% consisting of 4640 images, which is 2.8 times larger than the benchmark test data. Our proposed approach surpasses the first-place winner of the PIPAL NTIRE challenge by 3.4\% in SROCC and 2.7\% in PLCC. Therefore, our proposed model remains applicable to evaluate the quality of images generated by image restoration algorithms.

\noindent\textbf{Evaluation on individual distortions.} As shown in Table~\ref{dist_type_tid2013} and Table~\ref{dist_type_kadid}, we analyze the performance of our proposed model across different distortion types on the TID2013 and KADID-10k. Among the 24 distortion categories in TID2013, our model outperforms all others in 17 categories, exceeding the second best method by a margin of 13. Similarly, on KADID-10k, our approach exhibits superior performance across 19 out of 25 distortion types, surpassing the second best method by a margin of 15. These results validate our model can achieve stable performance across diverse distortions and show that retrieving instances with similar content and distortions can effectively represent the quality of the test image.
\subsection{Ablation study}
\begin{figure*}[tbp!]
	\centering
	\subfigure[ ]{%
		\begin{minipage}[t]{0.33\linewidth}
			\centering
			\includegraphics[width=\linewidth]{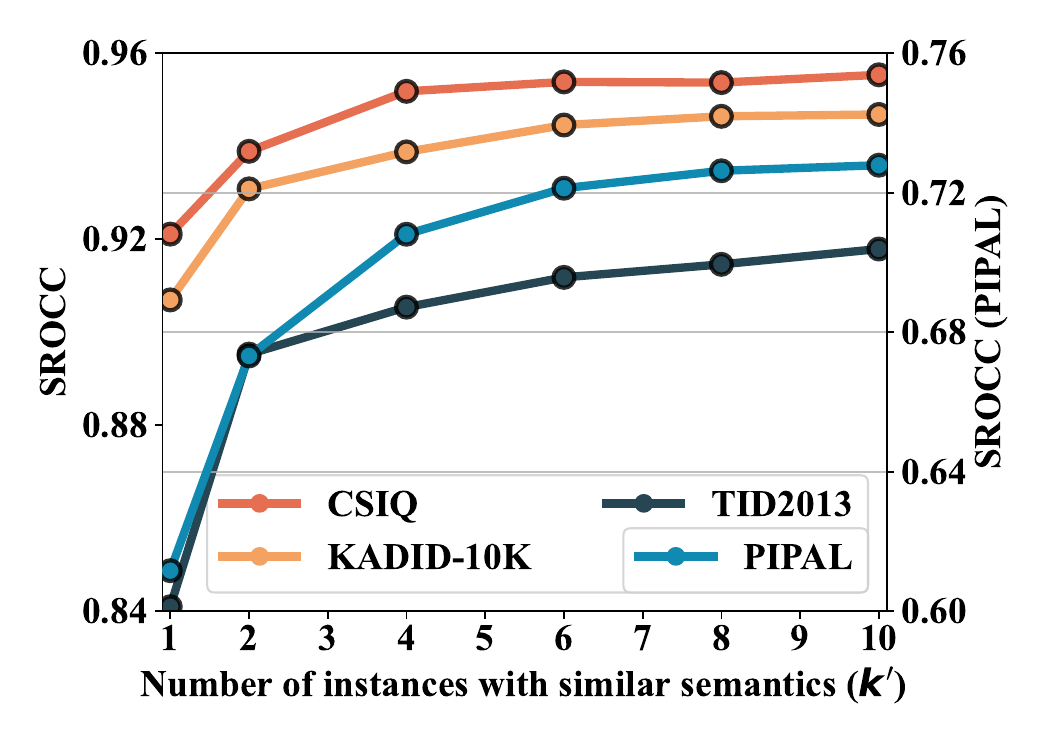}\noindent\vspace{-1mm}
			\label{fig:fig6-a}
		\end{minipage}%
	}%
	\subfigure[ ]{%
		\begin{minipage}[t]{0.33\linewidth}
			\centering
			\includegraphics[width=\linewidth]{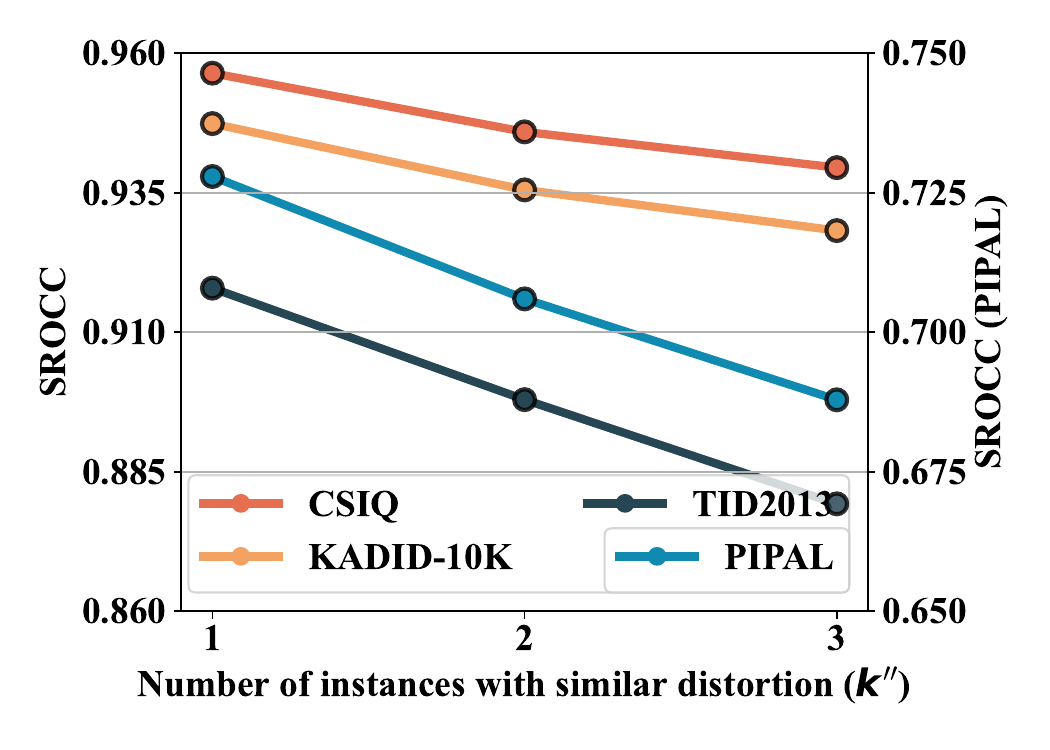}\vspace{-1mm}
			\label{fig:fig6-b}
		\end{minipage}%
	}%
	\subfigure[ ]{%
		\begin{minipage}[t]{0.33\linewidth}
			\centering
			\includegraphics[width=\linewidth]{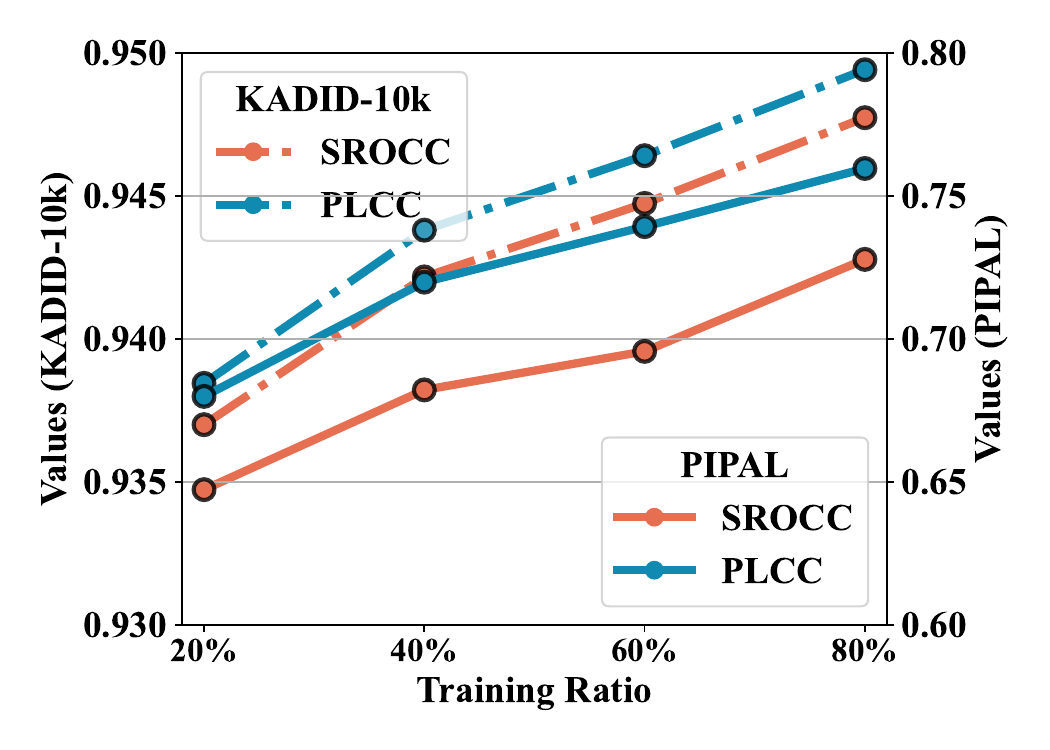}\vspace{-1mm}
			\label{fig:fig6-c}
		\end{minipage}%
	}%
	\centering
	\caption{Ablation experiments on the impact of different retrieval parameter settings on performance. (a) The number of instances with similar semantics, i.e., parameter $k'$. (b) The number of instances with similar distortions, i.e., parameter $k''$. (c) The impact of the number of instances to be retrieved on performance.}
	\label{fig:fig6}
\end{figure*}
\begin{figure}[!tp]
	\centering
	\includegraphics[scale=.5]{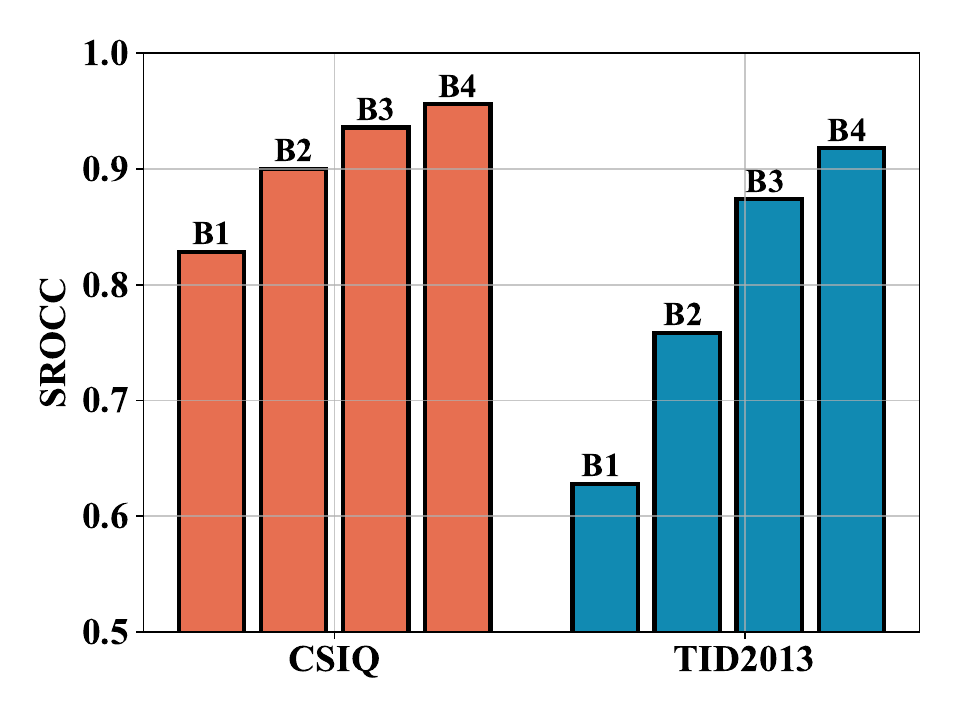}
	\caption{Ablation study on the performance of using features from different layers to retrieve instances. \textbf{B1} to \textbf{B4} denote features extracted from block 1 to 4 of the two modules.} 
	\label{fig_block}
\end{figure}

\noindent\textbf{Number of retrieved instances.}~As illustrated in Fig.~\ref{fig:fig6} (a) and (b), we analyze the impact of the number of retrieved instances (i.e, parameters $k'$ and $k''$) on performance. The results demonstrate that increasing $k'$ leads to performance improvements, especially when $k'$ is less than 6. This is because a limited number of retrieved instances can result in prediction bias. By increasing $k'$, the model is exposed to a richer set of instances, mitigating the bias issue. When $k'$ surpasses 6, the performance gains become minimal, implying that a modest number of similar instances effectively represents the quality of the test images. Fig.~\ref{fig:fig6} (b) shows that higher values of $k''$ lead to decreased performance. This reason is that the distorted similarity between retrieved instances and the test image is not consistent enough, resulting in a negative impact on predictions.
\begin{table}[tp!]
	\centering
	\caption{Ablation results on different aggregation strategies.}
	\fontsize{10.pt}{10.pt}\selectfont%
	\renewcommand{\arraystretch}{1.05}
	\resizebox{3.4 in}{!} 	{
		\begin{tabular}{cccccc}
			\toprule
			\multirow{2}{*}{\textbf{Number}} &\multirow{2}{*}{\textbf{Weighted}}& \multicolumn{2}{c}{\textbf{PIPAL}} & \multicolumn{2}{c}{\textbf{TID2013}} \\
			\cmidrule(r){3-4}\cmidrule(r){5-6}
			\textbf{$k'$}&\textbf{Average}& \textbf{SROCC$\uparrow$} & \textbf{PLCC$\uparrow$}& \textbf{SROCC$\uparrow$} & \textbf{PLCC$\uparrow$} \tabularnewline
			\midrule
			\multirow{2}{*}{\tabincell{c}{2}}
			
			&$\times$&0.6724 & 0.7069 & 0.8941& 0.8885 \tabularnewline
			&\checkmark&0.6732&  0.7075  &0.8953&0.8908\tabularnewline
			\midrule
			\multirow{2}{*}{\tabincell{c}{4}}
			
			&$\times$&0.7069& 0.7402&0.9091& 0.9089\tabularnewline
			&\checkmark& 0.7081&  0.7412& 0.9108&0.9106\tabularnewline
			\midrule
			\multirow{2}{*}{\tabincell{c}{6}}
			
			&$\times$&0.7204& 0.7503&0.9111&0.9183 \tabularnewline
			&\checkmark& 0.7213& 0.7515 & 0.9118&0.9192\tabularnewline
			\midrule
			\multirow{2}{*}{\tabincell{c}{8}}
			
			&$\times$&0.7256& 0.7566 &0.9137& 0.9202 \tabularnewline
			&\checkmark& 0.7263 &  0.7575 & 0.9146& 0.9226\tabularnewline
			\bottomrule
	\end{tabular}	}
	\label{tab_avg}
\end{table}
\begin{figure*}[!tp]
	\centering
	\includegraphics[scale=.52]{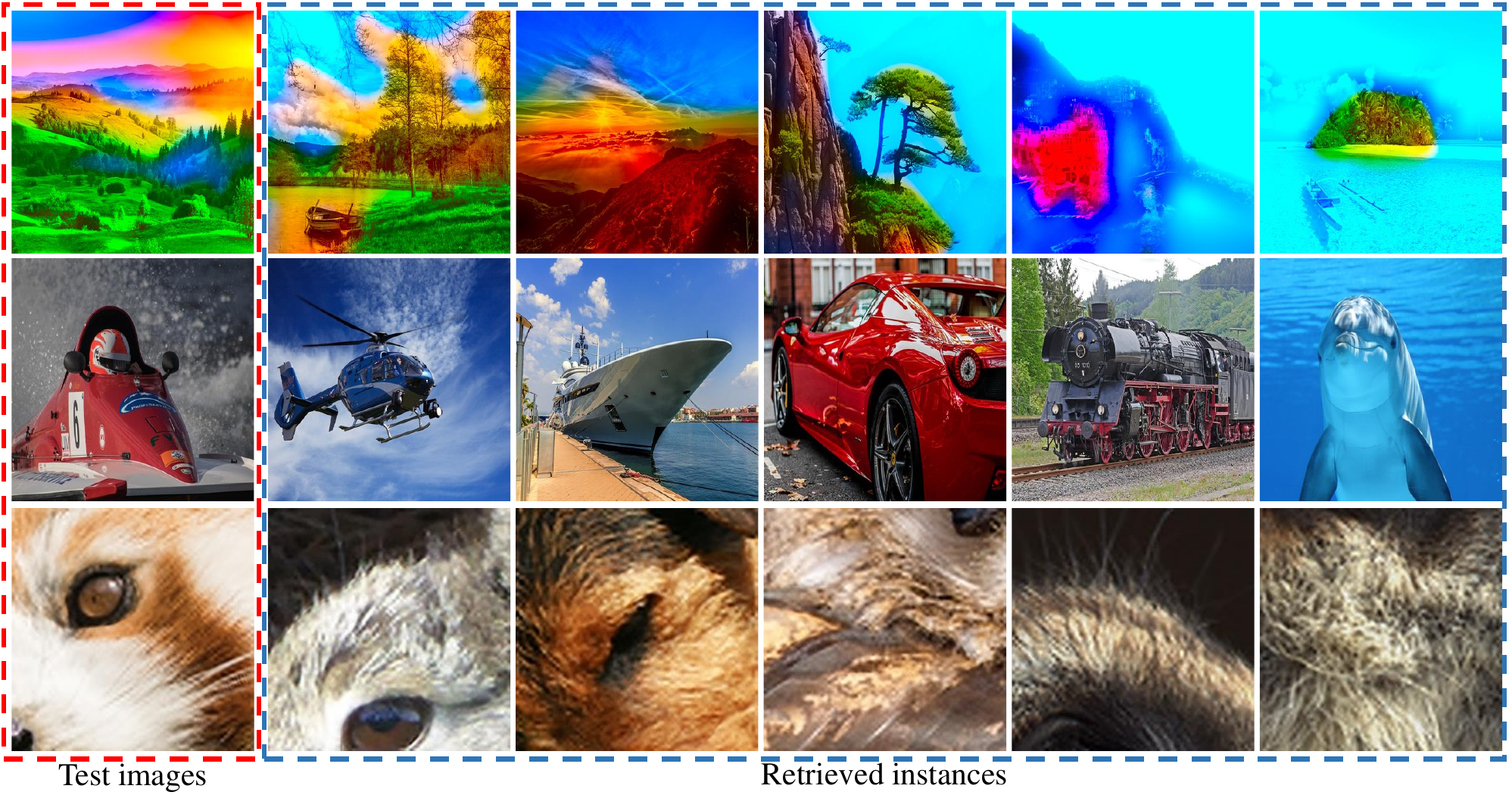} \vspace{-2mm}
	\caption{Visualization of retrieved instances of the proposed method.} 
	\label{fig_instances}
\end{figure*}

\noindent\textbf{Number of instances to be retrieved.}~We conducted ablation experiments to assess the impact of the number of instances to be retrieved. We fixed 20\% of the dataset and then randomly selected 20\%, 40\%, 60\%, and 80\% from the dataset for retrieval. Fig.~\ref{fig:fig6} (c) illustrates that increasing the number of instances enhances performance, as a larger set of instances can potentially contain examples that exhibit higher similarity to the test image. Notably, even with only 20\% of the training data, our proposed model still demonstrates good performance, achieving SROCC values of 0.937 on KADID-10k and 0.647 on PIPAL. These results outperform most of the BIQA methods listed in Table~\ref{single_dataset} and Table~\ref{pipal}.
\begin{table}[tp!]
	\centering
	\caption{Ablation results on different distance metrics.}
	\fontsize{8.pt}{8.pt}\selectfont%
	\renewcommand{\arraystretch}{1.05}
	\resizebox{3.4 in}{!} 	{
		\begin{tabular}{ccccc}
			\toprule
			\multirow{2}{*}{\textbf{Metric}} & \multicolumn{2}{c}{\textbf{CSIQ}} & \multicolumn{2}{c}{\textbf{TID2013}} \\
			\cmidrule(r){2-3}\cmidrule(r){4-5}
			\multirow{2}{*}{} & \textbf{SROCC$\uparrow$} & \textbf{PLCC$\uparrow$}& \textbf{SROCC$\uparrow$} & \textbf{PLCC$\uparrow$} \tabularnewline
			\midrule
			JS divergence&0.9495 & 0.9505 &0.9116 &0.9136 \tabularnewline
			Euclidean &0.9553& 0.9544&0.9189&  0.9238  \tabularnewline
			Manhattan &0.9546 &0.9542 & 0.9193&  0.9242   \tabularnewline
			Cosine&0.9562&  0.9540  &0.9179 & 0.9205  \tabularnewline
			\bottomrule
	\end{tabular}	}	
	\label{tab_distance}
\end{table}
\begin{table}[tp!]
	\centering
	\caption{Ablation results on different semantic models.}
	\fontsize{10.pt}{10.pt}\selectfont%
	\renewcommand{\arraystretch}{1.05}
	\resizebox{3.4 in}{!} 	{
		\begin{tabular}{ccccc}
			\toprule
			\multirow{2}{*}{\textbf{SC Module}} & \multicolumn{2}{c}{\textbf{CSIQ}} & \multicolumn{2}{c}{\textbf{TID2013}} \\
			\cmidrule(r){2-3}\cmidrule(r){4-5}
			\multirow{2}{*}{} & \textbf{SROCC$\uparrow$} & \textbf{PLCC$\uparrow$}& \textbf{SROCC$\uparrow$} & \textbf{PLCC$\uparrow$} \tabularnewline
			\midrule
			VGG-16& 0.9530 &0.9518 & 0.9078 &  0.9190 \tabularnewline
			VGG-19&0.9525 & 0.9528 & 0.9126 & 0.9210 \tabularnewline
			ResNet-18 &0.9562&  0.9540 &0.9179  &0.9205 \tabularnewline
			ResNet-34&0.9545& 0.9547 &0.9217& 0.9217 \tabularnewline
			ResNet-50 &0.9530 &0.9552 & 0.9117 & 0.9221 \tabularnewline
			\bottomrule
	\end{tabular}	}
	\label{tab_semantics}
\end{table}

\noindent\textbf{Features of different layers.}~We conducted ablation experiments by utilizing features from different layers for instance retrieval. Both the SC and DC modules consist of four blocks, and we extracted features from these four blocks to obtain representations of the image at different scales. Fig.~\ref{fig_block} shows that higher-level features exhibit superior performance. The reason is that higher-level features are more task-specific and provide content and distortion representations with higher discriminative power. In addition, they typically possess a larger receptive field, which helps to capture contextual information to recognize similar instances.

\noindent\textbf{Quality aggregation strategies.}~As shown in Table~\ref{tab_avg}, the results of the ablation experiments demonstrate a comparison between different quality aggregation strategies, including the simple average and the weighted average. The table demonstrates that, although the weighted average strategy exhibits slightly better performance compared to the simple average, the majority of the improvement in the model performance can be attributed to the effective similar instance retrieval scheme.

\begin{figure}[tbp!]
	\centering
	\subfigure[TID2013]{%
		\begin{minipage}[t]{0.49\linewidth}
			\centering
			\includegraphics[width=\linewidth]{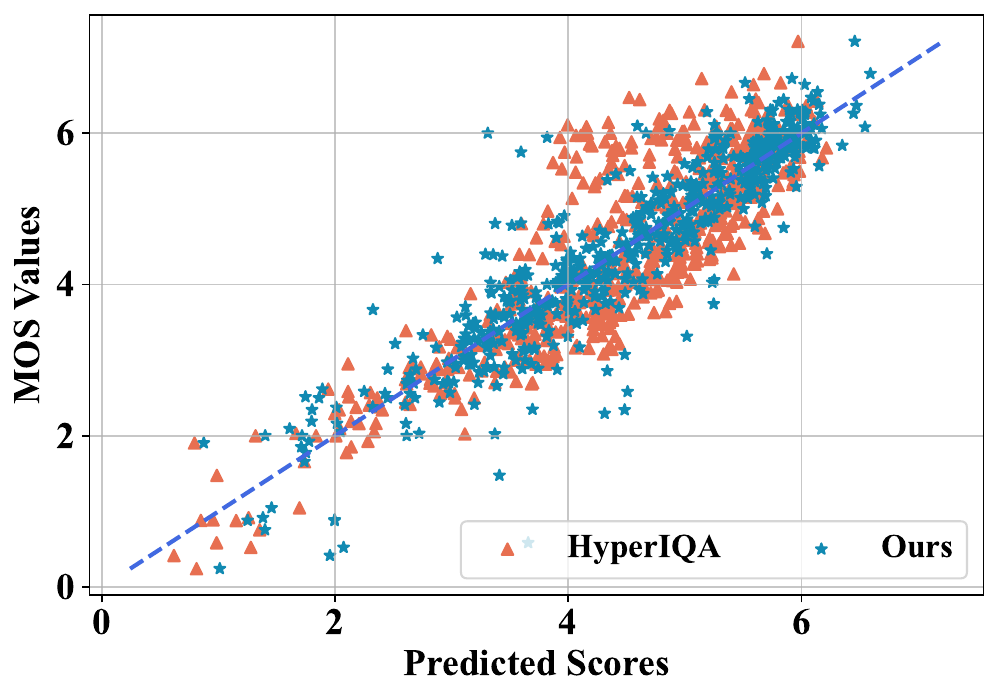}\noindent\vspace{-1mm}
		\end{minipage}%
	}%
	\subfigure[KADID-10k]{%
		\begin{minipage}[t]{0.49\linewidth}
			\centering
			\includegraphics[width=\linewidth]{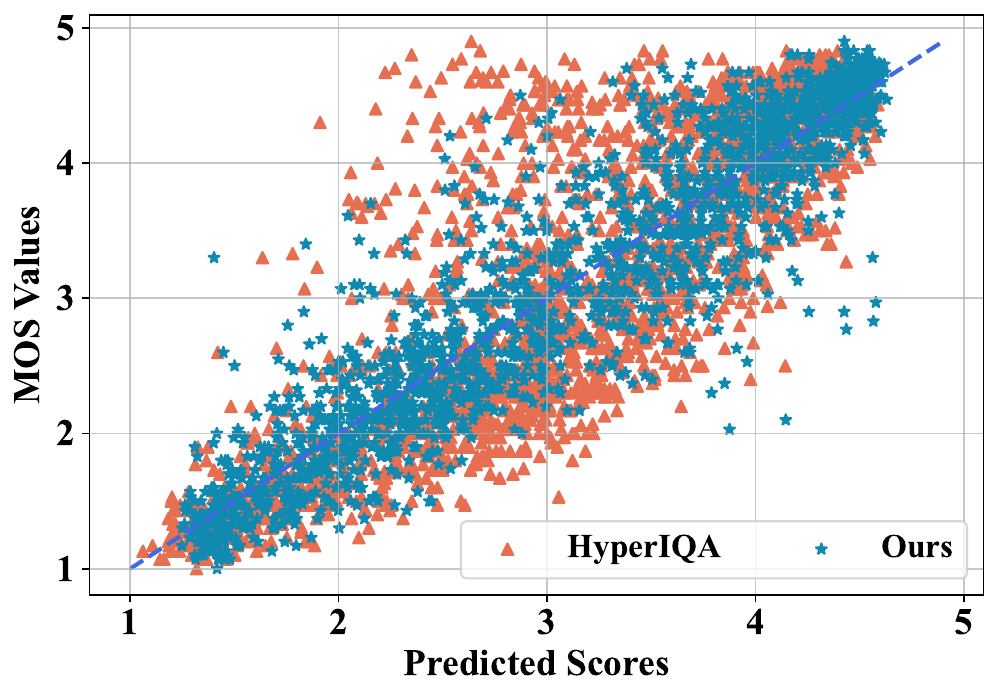}\vspace{-1mm}
		\end{minipage}%
	}%
	\caption{Correlation analysis between quality scores predicted by BIQA methods and the MOS values on TID2013 and KADID-10k.}
	\label{fig_preds}
\end{figure}

\noindent\textbf{Distance metrics.}Table~\ref{tab_distance} presents a comparison of different distance metrics, including Jensen-Shannon (JS) divergence, Euclidean distance, Manhattan distance, and cosine distance, for measuring semantic and distortion feature distances. JS divergence captures disparities in the distribution of classification layer logits. Employing feature-based metrics yields slightly better results than JS divergence. The results show that the method performs comparably on different feature distance metrics, highlighting its robustness to distance metric selection.

\noindent\textbf{Semantic models for SC module.} As shown in Table~\ref{tab_semantics}, the results of the ablation studies demonstrate the impact of different semantic models, e.g., VGG16~\cite{vgg} and ResNet50~\cite{resnet}, employed as the SC module. Despite variations in the number of parameters and classification accuracy on the ImageNet~\cite{imagenet}, all models demonstrate comparable performance, showing the independence of the proposed model from the precision of semantic representations generated by the SC module. This enhances the generalizability and flexibility of the proposed BIQA model.

\subsection{Discussion}
\begin{figure}[!tp]
	\centering
	\includegraphics[scale=.5]{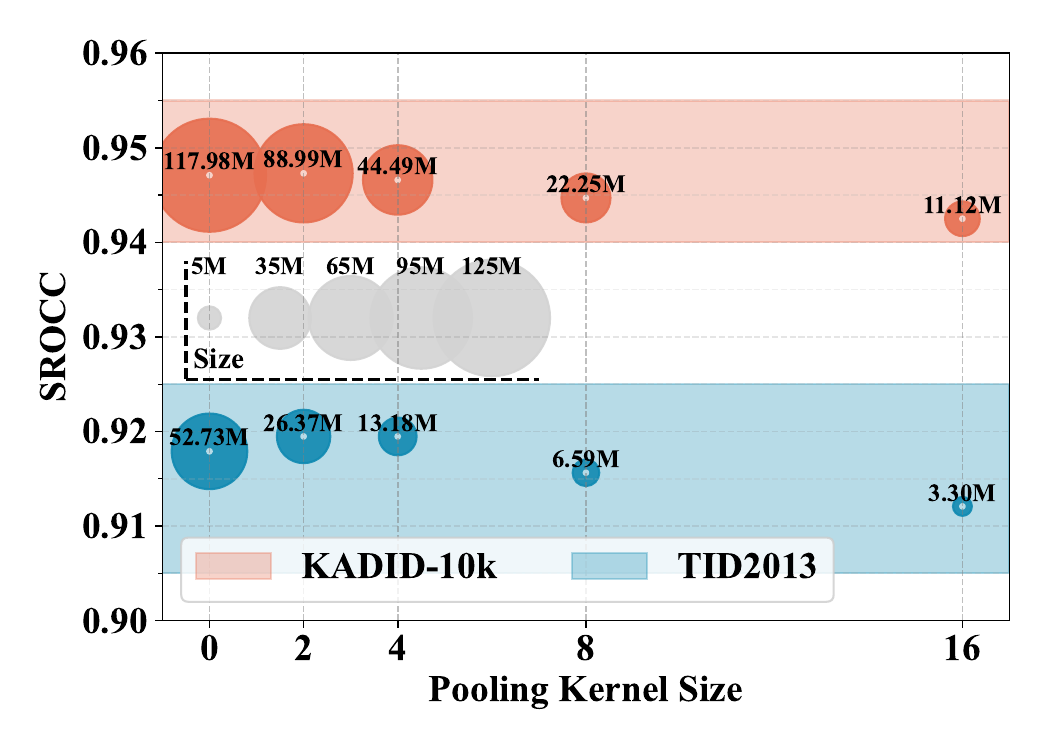}
	\caption{Ablation experiments on reduced dimensional features and performance.} 
	\label{fig_size}
\end{figure}
\noindent\textbf{Analysis of retrieval results.}~We present a visualization of the retrieved instances of our proposed model in Fig.~\ref{fig_instances}. The figure demonstrates the capability of our model to retrieve instances with similar content and distortion to the test images. For instance, as shown in the top row of the figure, our model is able to identify landscape pictures with severe color distortion that are similar to the test image. Furthermore, the model demonstrates effective detection of instances with similar contours and textures. In the second row of the figure, the retrieved instances exhibit similarity in terms of contours such as helicopters and ships. In the third row, the retrieved instances contain comparable animal fur textures to the query image. These results demonstrate the effectiveness of our model in perceiving image content and distortion and evaluating image quality through the retrieval of similar instances.

In addition, Fig.~\ref{fig_preds} (a) and (b) illustrate a correlation analysis between the quality scores predicted by the proposed method and MOS values on the TID2013 and KADID-10k, respectively. In comparison to HyperIQA~\cite{hyperiqa}, our approach exhibits a more concentrated distribution of predicted results around the diagonal line, signifying a higher level of consistency between the predicted values and human subjective perceptual judgments.

\noindent\textbf{Analysis of storage overhead.}~Considering practical application requirements, our model only needs to store image feature vectors and MOS values, rather than all retrieved images. To reduce storage needs, we apply one-dimensional max pooling for dimensionality reduction of these feature representations, using varying kernel sizes to balance compactness and performance impact. Fig.~\ref{fig_size} shows that small-scale dimensionality reduction can maintain model performance. This may be by reducing redundancy and noise to extract more discriminative features, while large-scale reduction can degrade retrieval accuracy through losing useful information. Still, relatively large-scale reduced features require minimal storage cost, without significant performance drops. For instance, with 16x feature reduction, TID2013 and KADID-10k require just 3.30MB and 11.12MB respectively, while preserving SROCC above 0.91 on TID2013 and 0.94 on KADID-10K. Thus, the proposed method can achieve competitive quality prediction performance with low storage requirements.

\section{Conclusion} \label{conclusion}
The existing regression-based BIQA models suffer from the presence of biased training samples. This issue can lead to a biased estimation of the model parameters, affecting its predictive ability. In this work, we have presented a novel regression-free approach for image quality assessment, which consists of two classification modules, namely the SC module and the DC module. The proposed model evaluates the quality of an input image by retrieving instances from a database with similar image content and distortion. Our method establishes local relationships between neighboring instances in the semantic and distortion feature spaces, rather than relying on the entire training set as in regression methods. This allows the proposed model to effectively reduce the prediction bias caused by a biased training set, by avoiding over-reliance on specific model parameters. Our experiments on seven benchmark IQA databases confirm the superiority of our method. Although not being trained on subjective quality scores, our approach achieves state-of-the-art performance in learnable distortion datasets and remains competitive with distortion unlearnable datasets.
\appendices

\ifCLASSOPTIONcaptionsoff

\fi
\bibliographystyle{IEEEtran}
\bibliography{reference}

\end{document}